\documentclass[11pt]{article}

\usepackage[margin=1in]{geometry}
\usepackage[T1]{fontenc}
\usepackage{lmodern}
\usepackage{amsmath,amssymb}
\usepackage{booktabs}
\usepackage{float}
\usepackage{microtype}
\usepackage{xcolor}
\usepackage[hidelinks]{hyperref}
\hypersetup{pdftitle={Blockwise Policy-Drift Gating for On-Policy Distillation},pdfauthor={Liwen Zheng and Haiyun Jiang}}

\title{Blockwise Policy-Drift Gating for On-Policy Distillation}
\author{Liwen Zheng\\Independent Researcher\and Haiyun Jiang\\Independent Researcher}
\date{June 2026}

\newcommand{\pold}{\pi_{\mathrm{old}}}
\newcommand{\ptheta}{\pi_{\theta}}

\newcommand{\sg}{\operatorname{stopgrad}}

\begin{document}
\maketitle

\begin{abstract}
On-policy distillation (OPD) trains a student policy using teacher signals
computed on trajectories sampled by the student itself.  Recent work shows
that sampled-token OPD can be fragile on long-horizon reasoning tasks and that
local teacher-support matching is a simple and effective repair.  This paper
introduces blockwise policy-drift gating, a lightweight student-only
old-current drift controller for OPD under rollout reuse.  The method computes
log-probability shifts between the behavior student and the current student on
the sampled token path, aggregates these shifts over fixed blocks or spans, and
uses the resulting detached, mean-normalized gates to reweight OPD position
losses.  It does not change teacher targets, teacher top-
\(K\) supports, or the rollout policy.  In a six-variant Qwen3 math reasoning benchmark with a uniform 200-step
training budget for all trained variants, we use pass@8 as the primary
problem-level solve-rate metric.  Fixed 64-token block gating improves
sampled-token OPD mean pass@8 from 0.4978 to 0.5160 across AIME24, AIME25,
MATH500, and AMC23.  On Teacher-TopK/LSM, Block64 gives the best
four-benchmark mean pass@8 among trained students.  The results identify local
old-current policy drift as a practical control signal for reused OPD rollouts
and motivate block-level gating as a simple default for improving solve-rate
robustness.
\end{abstract}

\section{Introduction}

On-policy distillation (OPD) is attractive for language-model post-training
because the teacher is queried on states actually visited by the student.  This
avoids training solely on static teacher traces and exposes the student to
feedback on its own rollouts.  However, long reasoning rollouts also make the
training signal fragile: a sampled token can be a narrow view of the teacher
distribution, and student-generated prefixes can move away from regions where
the teacher provides reliable local guidance.

Recent work on revisiting OPD identifies failure modes of sampled-token OPD and
proposes Teacher-TopK local support matching (LSM), which uses a truncated
teacher support rather than only the sampled token \cite{revisitingopd2026}.
Other recent studies examine OPD dynamics, teacher-student compatibility, and
broader OPD design choices \cite{rethinkingopd2026,opdsurvey2026}.  We build on
this line of work, but study a different source of mismatch.  In practical
PPO-style training, the same student rollout may be reused across multiple
optimization epochs.  During reuse, the current student can assign different
probabilities to the same response tokens than the behavior student that
generated them.  This old-current shift is not a teacher-reliability signal,
but it can indicate how stale a piece of rollout data is for the current
update.

We ask whether this sampled-path drift should modulate OPD losses and, if so,
at what granularity.  Token-level importance-ratio weights based on \(\exp(d_t)\) are a natural
baseline, but they can be noisy.  Sequence-level weighting is smoother but may be too
coarse for long reasoning traces.  We therefore focus on
a middle ground: blockwise and newline-delimited spanwise gates that aggregate
old-current log-probability shifts over local regions and broadcast the
resulting weight to positions in the region.

This paper makes three contributions:

\begin{enumerate}
    \item We formulate sampled-path old-current student drift as a detached
    plug-in position-level weight for existing OPD losses.
    \item We instantiate this rule as fixed-block and newline-span gates and
    apply the same gate family to sampled-token OPD and Teacher-TopK/LSM
    without changing teacher targets or teacher supports.
    \item We evaluate six method variants under a fixed 200-step OPD training
    budget on four math benchmarks, using pass@8 as the main problem-level
    solve-rate metric.  Fixed 64-token blocks improve sampled-token OPD solve
    rate and give the strongest aggregate pass@8 when added to
    Teacher-TopK/LSM.
\end{enumerate}

The method is best understood as a detached loss weighting rule: it changes
which reused rollout positions dominate optimization, not the teacher
supervision, the teacher support, or the rollout distribution.

\section{Background and Related Work}

\paragraph{Knowledge distillation and OPD.}
Knowledge distillation transfers behavior from a teacher model to a smaller or
more deployable student \cite{hinton2015distilling}.  For autoregressive
language models, off-policy distillation trains on fixed teacher-generated or
human-written traces and therefore suffers from a train-test distribution gap.
Generalized Knowledge Distillation (GKD) addresses this issue by training on
student-generated outputs and using teacher feedback on those outputs
\cite{agarwal2024opd}.  OPD follows the same central principle: the student
samples trajectories from its own policy, and the teacher provides dense
feedback on the prefixes that the student actually visits.  Recent OPD work for
reasoning models studies token-level supervision, sequence-level and hybrid
objectives, teacher-student compatibility, and broad design choices for OPD
systems \cite{rethinkingopd2026,opdsurvey2026}.

\paragraph{OPD failure modes and reliability controls.}
Long-horizon reasoning exposes several OPD-specific instability modes.
Revisiting OPD identifies brittle sampled-token supervision and proposes
Teacher-TopK local support matching (LSM), which replaces a one-token teacher
signal with a truncated teacher support at each visited prefix
\cite{revisitingopd2026}.  Stable-OPD studies length inflation and truncation
collapse and stabilizes training with a reference-based divergence constraint
and rollout mixture distillation \cite{stableopd2026}.  Entropy-aware OPD
changes the divergence when teacher entropy is high, addressing the diversity
loss and unstable signals induced by pure reverse-KL imitation
\cite{entropyopd2026}.  Horizon-control methods ask whether full rollouts are
necessary and show that progressive or truncated rollouts can improve OPD
efficiency while preserving strong supervision \cite{fullrollouts2026}.  Other
nearby work reallocates supervision within a rollout: Prune-OPD monitors local
teacher-student compatibility to downweight or truncate unreliable suffixes,
TIP studies token importance using student entropy and teacher-student
divergence, SCOPE routes rollouts through dual adaptive weighting paths,
SG-OPD gates distillation updates using verifier sign consistency, TrOPD
restricts or changes updates according to teacher-supervision reliability in
trust regions, and FiRe-OPD combines trajectory-level filtering with token-level
reweighting
\cite{pruneopd2026,tip2026,scopeopd2026,sgopd2026,tropd2026,fireopd2026}.
Among these, SCOPE and FiRe-OPD are closest in spirit because they also reweight
OPD supervision.  Our gate differs in both signal and granularity: it uses no
outcome label, verifier, teacher confidence, teacher--student divergence, or
trajectory filter.  Instead, it leaves the base OPD or LSM loss intact and
reweights positions using behavior-current student drift measured on the reused
sampled path.

\paragraph{Rollout reuse and freshness-aware OPD.}
PPO-style optimization commonly performs multiple minibatch or epoch updates
over data sampled by a behavior policy \cite{schulman2017ppo}.  Even in a
synchronous OPD pipeline, this creates a local old-current distinction during
rollout reuse.  Freshness-aware OPD is the closest neighboring line of work.
\(f\)-OPD studies stale rollouts in asynchronous or delayed-update settings,
decomposes objective discrepancy into rollout drift and supervision drift, and
introduces sample-level freshness control \cite{fopd2026}.  AsyncOPD studies
how stale data affects OPD under asynchronous execution and shows that KL
direction changes the stale-data problem: teacher-weighted forward KL is more
robust to stale rollouts, whereas student-weighted reverse KL is more
vulnerable \cite{asyncopd2026}.  Our setting is narrower and deliberately
orthogonal.  We do not design an asynchronous training system, a teacher-cache
strategy, or a sample refresh policy.  Instead, we keep the teacher supervision
fixed and study how old-current student log-probability shifts should be
aggregated before being used as detached weights on OPD token losses.

\paragraph{Sampled-token OPD and local support matching.}
Let \(x_i\) be a prompt and \(y_{i,1:T_i}\) a response sampled from a student
policy.  OPD constructs teacher-derived losses on visited states
\((x_i,y_{i,<t})\).  We write the base position-level OPD loss as
\(\ell^{\mathrm{base}}_{i,t}\), with response mask \(m_{i,t}\).  In our LSM
runs, the teacher top-\(K\) set, teacher probabilities, and normalization are
unchanged by policy-drift gating.  The gate is only an external weight on the
resulting position-level loss.

\section{Policy-Drift Gating}

For a sampled response token \(y_{i,t}\), define the old-current student
log-probability shift
\begin{equation}
 d_{i,t}
 = \log \ptheta(y_{i,t}\mid x_i,y_{i,<t})
 - \log \pold(y_{i,t}\mid x_i,y_{i,<t}).
\end{equation}
We use ``policy drift'' as shorthand for this sampled-path old-current shift;
it is not a full distributional KL estimate.  Both log-probabilities are
computed on the same sampled response tokens with the same response mask and
causal alignment.  In the implementation used here, \(\log \pold\) is
recomputed by the actor before policy update, so the ratio is based on raw
actor log-probabilities rather than top-\(p\)-renormalized rollout
probabilities.  The teacher model is not used in this ratio.

\paragraph{Gate shape.}
For an aggregated score \(s\), we use a detached soft drift gate
\begin{equation}
 g = \exp(-\tau |s|),
\end{equation}
with \(\tau=1\).  Gates are mean-normalized over valid response tokens:
\begin{equation}
 \bar g =
 \frac{\sum_{i,t} m_{i,t} g_{i,t}}{\sum_{i,t}m_{i,t}+\epsilon},
 \qquad
 \tilde g_{i,t}=\sg\left(\frac{g_{i,t}}{\bar g+\epsilon}\right).
\end{equation}
The normalized gate can be larger than one.  Its purpose is to preserve the
average loss scale over valid response tokens while changing the relative
emphasis of positions with different old-current shifts.

\paragraph{Granularity.}
The main matrix evaluates two aggregation granularities:

\begin{itemize}
    \item \textbf{Block64-Gate:} split responses into fixed 64-token blocks,
    average \(d_{i,t}\) inside each block, and broadcast the gate to tokens in
    the block.
    \item \textbf{NewlineSpan-Gate:} split responses using newline delimiters,
    merge short spans, split long spans, and fall back to fixed chunks when no
    delimiter is present.
\end{itemize}

The final weighted objective is
\begin{equation}
\mathcal{L}_{\mathrm{gate}}
=
\frac{\sum_{i,t} m_{i,t}\tilde g_{i,t}\ell^{\mathrm{base}}_{i,t}}
{\sum_{i,t} m_{i,t} + \epsilon}.
\end{equation}
The denominator remains the original response-mask denominator.  Detachment
prevents the gate from becoming an additional objective on the current policy;
it only controls how much each base OPD position loss contributes.

\section{Experiments}

\paragraph{Models and codebase.}
We use Qwen3-1.7B-Base as the student and Qwen3-4B-Base-GRPO as the teacher,
following the Qwen3 model family \cite{qwen32025}.  The teacher checkpoint is a
post-RL Qwen3-4B-Base checkpoint trained with GRPO for mathematical reasoning
\cite{qwen3grpo2026}.  The implementation is based on the public
\texttt{revisiting\_opd} codebase with local policy-drift-gating
modifications.

\paragraph{Training data and setup.}
Training prompts are drawn from the public DAPO-Math-17k math corpus
\cite{dapomath17k}.  We use the released parquet file as a prompt-and-answer
pool and convert each row into the math environment format used by
\texttt{revisiting\_opd}: a single user-turn prompt, a rule-based ground-truth
answer field, and per-example metadata carrying the normalized question hash.
The prepared training table used in our runs contains 1,791,700 math rows;
this is a row count from the released parquet table, not a count of unique
prompts.  The preprocessing script also checks exact normalized-question hashes
against the prepared evaluation files; no training rows are removed by this
exact-hash filter in our runs.  This guard should not be read as semantic
decontamination.  We do not add separate teacher-written solution traces during
preprocessing.  Teacher signals are computed online on the student-sampled
rollout prefixes during OPD training.

We use a fixed 200-step training budget for every trained variant.  This is a
uniform benchmark budget for controlled method comparison, not a per-method
early-stopping selection.  The choice is consistent with recent OPD reliability
benchmarking practice, where TrOPD also trains OPD baselines for 200 steps under
shared settings \cite{tropd2026}.  Our training configuration uses maximum
prompt length 2048, maximum response length 16384, rollout group size 8,
top-\(p=0.9\), temperature 1.0, Teacher-TopK \(K=32\) for LSM rows, learning
rate \(2\times 10^{-6}\), and two PPO epochs.  Each individual training run uses
A800 80GB GPU\@.  We use a train batch size of 4 for memory and wall-clock
feasibility while preserving the rollout group size and response length.

\paragraph{Evaluation.}
All trained-student rows are external vLLM evaluations of the resulting 200-step models.
We sample \(n=8\) generations per problem with temperature 1.0, top-\(p=0.9\),
and max new tokens 16384.  Pass@8 is our primary metric: a problem is counted
as solved if any of the eight generations is correct.  We also compute avg@8,
the mean sample accuracy over the eight generations, as an auxiliary diagnostic
but do not use it to rank the main variants.  We evaluate on MATH500
\cite{math500}, AIME 2024 \cite{aopsaime2024}, MathArena AIME 2025
\cite{matharena2026}, and AMC23.  The AMC23 file contains 40 problems from the
2023 AMC 12A/12B sets and stores an Art of Problem Solving source URL for each
item \cite{aopsamc2023}.  We report unweighted mean pass@8 across these four
sets as the primary aggregate.

\paragraph{Main matrix.}
The six trained students are:

\begin{center}
\begin{tabular}{lll}
\toprule
Method & Base loss & Gate \\
\midrule
OPD & sampled-token OPD & none \\
OPD + Block64 & sampled-token OPD & fixed 64-token block gate \\
OPD + NewlineSpan & sampled-token OPD & newline-delimited span gate \\
LSM & Teacher-TopK & none \\
LSM + Block64 & Teacher-TopK & fixed 64-token block gate \\
LSM + NewlineSpan & Teacher-TopK & newline-delimited span gate \\
\bottomrule
\end{tabular}
\end{center}

\section{Results}

\textbf{Main solve-rate comparison.}
Table~\ref{tab:main} reports pass@8, the primary problem-level metric in this
study.  All trained students substantially improve over the base student.
Among the trained 1.7B students, LSM + Block64 obtains the best mean pass@8,
and Block64 improves sampled-token OPD from 49.8 to 51.6.  The gains are not
uniform across every benchmark, but the aggregate pattern is favorable to
blockwise gating: Block64 improves OPD directly and gives the strongest LSM
solve-rate aggregate.

\begin{table}[H]
\centering
\small
\setlength{\tabcolsep}{6pt}
\begin{tabular}{lccccc}
\toprule
Method & MATH500 & AIME24 & AIME25 & AMC23 & Avg. \\
\midrule
Base student & 77.2 & 13.3 & 10.0 & 60.0 & 40.1 \\
Sampled-token OPD & 85.8 & 20.0 & 13.3 & \textbf{80.0} & 49.8 \\
\textbf{\quad + Block64 gate} & \textbf{86.4} & 20.0 & 20.0 & \textbf{80.0} & 51.6 \\
\quad + NewlineSpan gate & 86.0 & \textbf{26.7} & 16.7 & 75.0 & 51.1 \\
Teacher-TopK/LSM & 86.0 & 23.3 & 20.0 & 70.0 & 49.8 \\
\textbf{\quad + Block64 gate} & \textbf{86.4} & 20.0 & \textbf{26.7} & \textbf{80.0} & \textbf{53.3} \\
\quad + NewlineSpan gate & 86.2 & 20.0 & \textbf{26.7} & \textbf{80.0} & 53.2 \\
Teacher reference & 90.8 & 46.7 & 33.3 & 90.0 & 65.2 \\
\bottomrule
\end{tabular}
\caption{Main pass@8 solve-rate results under the fixed 200-step OPD training budget.  Scores are percentages.  Avg. is the
unweighted mean over MATH500, AIME24, AIME25, and AMC23, computed from
unrounded evaluation outputs; displayed benchmark scores are rounded to one
decimal.  Bold numeric values mark the best trained 1.7B student for each
column, with ties retained.  The teacher is a larger 4B reference model and is
not a size-matched comparison.}
\label{tab:main}
\end{table}

Auxiliary avg@8 diagnostics are used only as a sample-level sanity check.  On
the sampled-token OPD base loss, they are aligned with the main table: Block64
improves mean avg@8 from 0.3157 to 0.3231 while also improving pass@8.  Since the target of this paper is solve-rate robustness under rollout reuse,
we use pass@8 for the main ranking.

Newline-delimited spans are competitive but not uniformly reliable.  Under the fixed 200-step budget,
NewlineSpan is close to Block64 on aggregate pass@8, especially when
combined with LSM, but it does not improve sampled-token OPD as consistently as
Block64.  A plausible interpretation is that delimiter spans are sensitive to
the model's formatting behavior and to the underlying loss.  We therefore
treat newline span gating as a useful heuristic probe, not as a semantic
segmentation method.

\section{Discussion}

The results fit the intended scope of the method.  Policy-drift gating is not a
replacement for teacher-support fixes such as LSM, and it is not a full stale-rollout correction framework.  Instead, it is an orthogonal weighting mechanism
that asks whether reused rollout tokens should contribute equally when the
current student has moved away from the behavior student.  Fixed blocks offer a
simple compromise between noisy token weights and overly coarse sequence
weights.  The empirical advantage of Block64 on sampled-token OPD and on LSM
pass@8 supports block-level drift gating as a practical default for this
rollout-reuse setting.  This is a solve-rate robustness result rather than a
claim that any single gate dominates every secondary metric or training budget.

A plausible explanation for the Block64 advantage is that old-current student
drift is locally coherent but tokenwise noisy.  Individual token log-ratio
shifts can be dominated by lexical, formatting, or tokenizer-level effects,
whereas response-level averages wash out stale regions in long reasoning
traces.  Fixed 64-token blocks provide a middle scale: they smooth token-level
spikes while preserving enough locality to identify reasoning segments whose
behavior-policy likelihood has changed under rollout reuse.  The auxiliary
avg@8/pass@8 comparison further suggests that blockwise gating mainly improves
sample quality or solve-rate robustness rather than uniformly expanding
rare-sample coverage, motivating softer or diversity-preserving gates in
future work.

The distinction from freshness-aware asynchronous OPD is important.  \(f\)-OPD
uses sample-level freshness control to regulate stale samples in asynchronous
training \cite{fopd2026}.  Our experiments do not test asynchronous execution,
sample buffering, or refresh policies.  They instead isolate a smaller question
inside synchronous rollout reuse: whether old-current student log-probability
shifts, aggregated locally along the sampled response path, are useful as loss
weights for OPD token losses.

The teacher reference remains much stronger than all trained students.  This is
expected: the teacher is a larger 4B GRPO model, while all trained students are
1.7B models.  The teacher row is included to contextualize the distillation
problem, not as a size-matched baseline.

\section{Limitations}

This is a preliminary empirical study.  The matrix uses one student model, one
larger teacher, one math training setup, and no repeated random seeds.  AIME-style sets have small sample counts, so differences on AIME25 should be
interpreted cautiously.  The current matrix also does not include granularity and ratio-weighting
controls such as Token-Gate, a clipped token-level importance-ratio baseline
(Token-IS), or Sequence-Gate.  Without those controls, we should not claim
that 64-token blocks are universally optimal.

The gate score for a block averages signed log-probability shifts.  Opposite-sign token shifts can therefore cancel within a block, even when the magnitude
of local old-current movement is large.  Mean-absolute or RMS log-ratio block
gates are natural controls for future work.  This v1 also focuses on downstream
benchmark results rather than gate-distribution diagnostics such as log-ratio
percentiles, effective token ratios, or gradient-norm tails.  Finally, the
method is a policy-drift-aware weighting rule, not an unbiased off-policy
estimator.

\section{Conclusion}

We introduced sampled-path policy-drift gating as a simple plug-in weight for
OPD and Teacher-TopK/LSM losses.  In a controlled fixed-budget 200-step math reasoning study, fixed 64-token
block gating improves sampled-token OPD on the primary pass@8 metric and gives
the strongest four-benchmark mean pass@8 when added to LSM.  Auxiliary avg@8 diagnostics support the OPD-side gain
and are kept as a secondary sample-level check.  These results show that
old-current student drift is a useful signal for OPD under rollout reuse, and
that the granularity of this signal should be evaluated primarily through
problem-level solve rate.

\paragraph{Next steps.}
For the next version, we plan to expand the controlled comparison around the
same OPD objective.  We will add token-, sequence-, and importance-ratio
baselines: Token-Gate applies the drift penalty at individual-token granularity,
Token-IS uses a detached clipped token-level ratio \(\exp(d_t)\) as a
conventional importance-weighting control, and Sequence-Gate applies one drift
weight to the whole response.  We will also sweep block sizes such as 32, 64,
and 128 tokens, test alternatives to signed block means, including
mean-absolute, RMS, and asymmetric drift statistics, and evaluate residual or
clipped normalization to avoid over-suppressing rare but useful samples.  Finally, we will add log-ratio percentile,
effective-token-ratio, and gradient-norm-tail diagnostics, then repeat the study
across additional model sizes, datasets, seeds, and training budgets to evaluate
compatibility with teacher-reliability and verifier-based OPD controls.

\end{document}